\documentclass[11pt]{article}
\usepackage{hyperref}
\usepackage[pdftex]{graphicx}
\usepackage{subfig}
\usepackage{colortbl}
\usepackage{wrapfig}
\usepackage{sidecap}
\usepackage[algo2e]{algorithm2e}
\usepackage{pifont}
\usepackage[export]{adjustbox}
\usepackage{coling2020}
\usepackage{times}
\usepackage{latexsym}
\usepackage{url}
\usepackage{graphicx}
\usepackage{amsmath}
\usepackage{amsfonts}
\usepackage{booktabs}
\usepackage{multicol}
\usepackage{multirow}
\usepackage{tabularx}
\usepackage{color}
\usepackage{lineno}
\usepackage{mathtools}
\usepackage{epsfig}
\usepackage{amssymb}
\usepackage{algorithm}
\usepackage{algorithmicx}

\usepackage{enumerate}

\definecolor{gray}{RGB}{87, 87, 87}
\definecolor{red}{RGB}{173, 35, 35}
\definecolor{blue}{RGB}{42, 75, 215}
\definecolor{green}{RGB}{29, 105, 20}
\definecolor{brown}{RGB}{129, 74, 25}
\definecolor{purple}{RGB}{129, 38, 192}
\definecolor{cyan}{RGB}{41, 208, 208}
\definecolor{yellow}{RGB}{189, 167, 0}
\definecolor{Red}{rgb}{0.68, 0.05, 0.0}
\definecolor{Blue}{rgb}{0.0, 0.0, 0.61}
\definecolor{Blue1}{RGB}{214, 235, 245}
\definecolor{Blue2}{RGB}{235, 245, 250}
\definecolor{lime}{RGB}{60,179,113}

\colingfinalcopy 

\title{Variational Reward Estimator Bottleneck: Learning Robust Reward Estimator for Multi-Domain Task-Oriented Dialog}

\author{Jeiyoon Park$^1$, Chanhee Lee$^1$, Kuekyeng Kim$^{1,2}$, Heuiseok Lim$^1$ \\
    $^1$Korea University, Seoul, Republic of Korea \\
    {\tt \{k4ke, chanhee0222, limhseok\}@korea.ac.kr} \\ 
    $^2$Massachusetts Institute of Technology \\
    {\tt kuekyeng@media.mit.edu} \\
    }
\date{}
\begin{document}
\maketitle
\begin{abstract}
  Despite its notable success in adversarial learning approaches to multi-domain task-oriented dialog system, training the dialog policy via adversarial inverse reinforcement learning often fails to balance the performance of the policy generator and reward estimator. During optimization, the reward estimator often overwhelms the policy generator and produces excessively uninformative gradients. We proposes the Variational Reward estimator Bottleneck (VRB), which is an effective regularization method that aims to constrain unproductive information flows between inputs and the reward estimator. The VRB focuses on capturing discriminative features, by exploiting information bottleneck on mutual information. Empirical results on a multi-domain task-oriented dialog dataset demonstrate that the VRB significantly outperforms previous methods.
\end{abstract}
\section{Introduction}
\begin{wrapfigure}[16]{r}{0.36\textwidth}
{
\centering
\vspace{-2mm}
\includegraphics[width=0.97\linewidth]{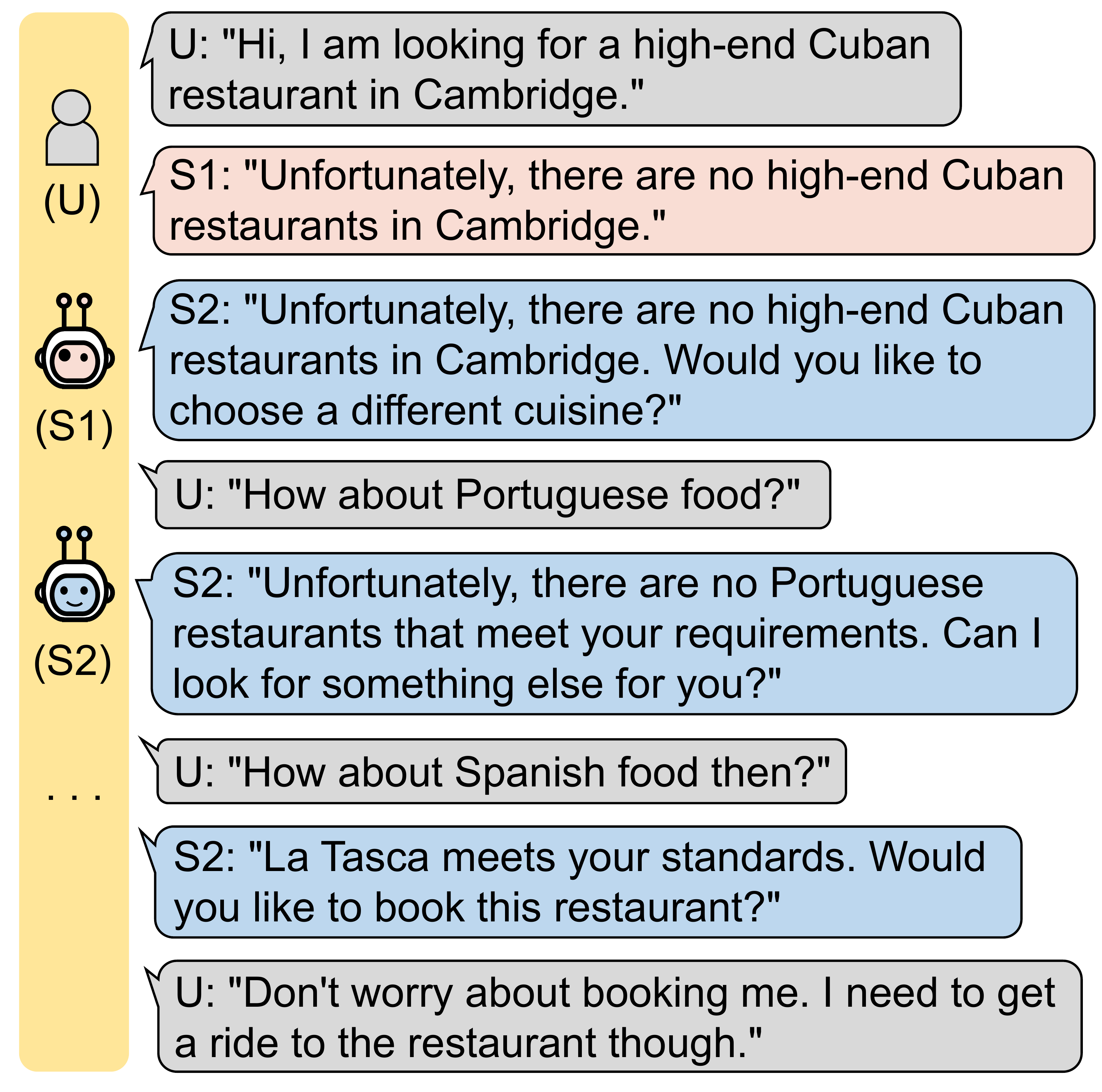}
}
\footnotesize
\vspace*{-2mm}
\caption{The system (S2) that uses well-specified rewards can guide the user through the goal while S1 can't.} 
\label{fig:1}
\end{wrapfigure}

While deep reinforcement learning (RL) have emerged as a promising solution for complex and high-dimensional decision-making problems, the determination of an effective reward function remains a challenge, especially in multi-domain task-oriented dialog systems. Many recent works have struggled on sparse-reward environments and employed a handcrafted reward function as a breakthrough \cite{zhao-eskenazi-2016-towards,Dhingra_2017,Shi_2018,shah-etal-2018-bootstrapping}. However, such approaches are often unable to guide the dialog policy through user goals. For instance, as illustrated in Figure \ref{fig:1}, the user can't reach the goal because the system (S1) that exploits the handcrafted rewards completes the dialog session too early. Moreover, the user goal usually varies as the dialog proceeds. 

Inverse Reinforcement Learning (IRL) \cite{russell1998learning,article} and MaxEnt-IRL \cite{ziebart2008maximum} tackles the problem of recovering reward function and using this reward function to generate optimal behavior. Although Generative adversarial imitation learning (GAIL) \cite{NIPS2016_6391}, which exploits the GANs framework \cite{NIPS2014_5423}, has proven that the discriminator  
can be defined as a reward function, GAIL fails to generalize and recover the reward function. Adversarial inverse reinforcement learning (AIRL) \cite{DBLP:conf/iclr/FuLL18} enables GAIL to take advantage of disentangled rewards. Guided dialog policy learning (GDPL) \cite{takanobu-etal-2019-guided} uses AIRL framework to construct the reward estimator for multi-domain task-oriented dialogs. However, these methods often encounter difficulties in balancing the performance of the policy generator and reward estimator, and produce excessively uninformative gradients.   

In this paper, we propose the Variational Reward Estimator Bottleneck (VRB), an effective regularization algorithm. The VRB uses information bottleneck \cite{tishby99information,alemi2016variational,peng2018variational} to constrain unproductive information flows between dialog state-action pairs and internal representations of the reward estimator, thereby ensuring highly informative gradients and robustness. The experiments demonstrate that the VRB achieves the state-of-the-art performances on a multi-domain task-oriented dataset.
\clearpage
\section{Background}
\label{Background}
\subsection{Dialog State Tracker And User Simulator}
The dialog state tracker (DST) \cite{Wu_2019}, which takes dialog action $a$ and dialog history as input, updates the dialog state $x$ and belief state $b$ for each slot\footnote{For background and notations on MDP, see Appendix \ref{app a.1}.}. For example, in Figure \ref{fig:2}, DST observes the  user goal where the user wishes to go. At dialog turn t, the dialog action is represented as a slot and value pair (\textit{e.g. Attraction: (area, centre), (type, concert hall)}). Given the dialog action, DST encodes the dialog state as $x_{t} = [a^{u}_t; a_{t-1}; b_{t}; q_t]$. The user simulator $\mu(a^{u},t^{u}|x^u)$ \cite{schatzmann-etal-2007-agenda,gur2018user} extracts the dialog action $a^u$ corresponding to the dialog state $x^u$. $t^u$ stands for whether user goal is achieved during conversation. Note that the DST and the user simulator can't achieve the user goal without well-defined reward estimation.
\subsection{Reward Estimator}
The reward estimator \cite{takanobu-etal-2019-guided}, which is a core component in multi-domain task-oriented dialog systems, evaluates dialog state-action pairs at dialog turn $t$ and estimates the reward that is used for guiding the dialog policy through the user goal. Based on MaxEnt-IRL \cite{ziebart2008maximum}, each dialog session $\tau$ in a set of human dialog sessions $\mathcal{D} = \{\tau_1,\tau_2, ..., \tau_H\}$ can be modeled as a Boltzmann distribution that does not exhibit additional preferences for any dialog sessions: $f_{\zeta}(\tau) = \log{\left( \cfrac{\exp(\mathcal{R_{\zeta})}}{Z} \right) }$ where $\mathcal{R}_{\zeta} = \sum_{t=0}^T\gamma^{t}r_{\zeta}(x_t,a_t)$, $Z$ is a partition function, $\zeta$ is a parameter of reward function, and $\mathcal{R}_{\zeta}$ denotes a discounted cumulative reward. The reward estimator can be trained using gradient-based optimization as follows (for the complete derivation, see Appendix \ref{app a.2}):
\begin{equation}
\begin{aligned}
\label{Equation.3}
{L_{f}(\zeta,\psi) = \mathbb{E}_{\tau \sim \mathcal{D}}[f_{\zeta,\psi}(x_t, a_t, x_{t+1})] - \mathbb{E}_{\tau \sim \pi}[f_{\zeta,\psi}(x_t, a_t, x_{t+1})]}
\end{aligned}
\end{equation}
\subsection{Policy Generator}
The policy generator \cite{pmlr-v37-schulman15,schulman2017proximal} encourages the dialog policy $\pi_\theta$ to determine the next action that maximizes the reward function $\hat{r}_{\zeta,\psi}(x_t,a_t,x_{t+1}) = f_{\zeta, \psi}(x_t,a_t,x_{t+1}) - \log{\pi}_{\theta}(a_t | x_t)$ (the full derivation is available in Appendix \ref{app a.3}):
\begin{equation}
\begin{gathered}
\label{Equation.4}
L^{CLIP}_\pi(\theta) = \mathbb{E}_{x,a \sim \pi}[\min(\xi_t(\theta) \hat{A}_t, \, \text{clip}(\xi_t(\theta), 1 - \epsilon, 1 + \epsilon)\hat{A}_t)] \\ L^{VF}_t(\theta) = -\left( V_{\theta} \,-\, \sum^{T}_{k=t}\gamma^{k-t}\hat{r}_k \right)^2 
\end{gathered}
\end{equation}
where $\hat{A}_t = \delta_t + \gamma \lambda \hat{A}_{t+1}$, $\delta_t = \hat{r}_{\zeta, \psi} + \gamma V(x_{t+1}) - V(x_t)$, and $\delta$ is the TD residual \cite{schulman2015high}. $\xi_t(\theta) = \frac{\pi_\theta(a_t|x_t)}{\pi_{\theta_\text{old}}(a_t|x_t)}$ and $V_\theta$ is the state-value function. Epsilon and $\lambda$ are hyper-parameters. 
\section{Variational Reward Estimator Bottleneck}
The Variational information bottleneck \cite{tishby99information,alemi2016variational,peng2018variational} is an information-theoretic approach that restricts unproductive information flow between inputs and the discriminator. Inspired by this concept, we propose a regularized objective that constrains the mutual information between encoded state-action pairs and original inputs, thereby ensuring highly informative internal representations and robust adversarial model. Our proposed method learns an encoder that is maximally informative regarding human dialogs. To this end, we employ a stochastic encoder and an upper bound constraint on the mutual information between the dialog states $X$ and latent variables $\mathbf{Z}$:
\begin{equation}
\begin{gathered}
\label{Equation.5}
L_{f,\mathbf{E}}(\zeta,\psi) = \mathbb{E}_{x,a \sim \mathcal{D}} [\, \mathbb{E}_{\mathbf{z} \sim \mathbf{E}(\mathbf{z}|x_t,x_{t+1})}[\, f_{\zeta,\psi}(\mathbf{z}_g, \mathbf{z'}_h, \mathbf{z}_h)]\,]\, - \, \mathbb{E}_{x,a \sim \pi} [\, \mathbb{E}_{\mathbf{z} \sim \mathbf{E}(\mathbf{z}|x_t,x_{t+1})}[\, f_{\zeta,\psi}(\mathbf{z}_g, \mathbf{z'}_h, \mathbf{z}_h)]\,]
\\ \text{s.t.} \quad I(Z,X) \le I_c
\end{gathered}
\end{equation}
where $f_{\zeta,\psi}(\mathbf{z}_g, \mathbf{z}'_h, \mathbf{z}_h) = D_g(\mathbf{z}_g) + \gamma D_{h}(\mathbf{z'}_h) + D_h(\mathbf{z}_h)$ and $D$ is modeled with nonlinear function.
Note that $f_{\zeta,\psi}(\mathbf{z}_g, \mathbf{z}'_h, \mathbf{z}_h)$ is divided into the three terms $D_g(\mathbf{z}_g)$, $\gamma D_{h}(\mathbf{z'}_h)$, and $D_h(\mathbf{z}_h)$, based on GANs \cite{NIPS2014_5423}, GAN-GCL \cite{10.5555/3045390.3045397}, and AIRL \cite{DBLP:conf/iclr/FuLL18}. $D_g$ represents the encoded disentangled reward approximator with the parameter $\zeta$, and $D_{h}$ is the encoded shaping term with the parameter $\psi$. Stochastic encoder $\mathbf{E}(\mathbf{z}|x_t,x_{t+1})$ can be defined as $\mathbf{E}(\mathbf{z}|x_t,x_{t+1}) = \mathbf{E}_{g}(\mathbf{z}_g|x_t) \cdot \mathbf{E}_{h}(\mathbf{z}_h|x_t) \cdot \mathbf{E}_{h}(\mathbf{z'}_h|x_{t+1})$ which maps states to a latent distribution $\mathbf{z}$: $\mathbf{E}(\mathbf{z}|x_t) = \mathcal{N}(\mu_{\mathbf{E}}(x_t), \Sigma_{\mathbf{E}}(x_t))$.  $r(\mathbf{z}) = \mathcal{N}(0, I)$ is standard gaussian and $I_c$ stands for an enforced upper bound on mutual information. To optimize $L_{f,\mathbf{E}}(\zeta,\psi)$, VRB introduces a Lagrange multiplier $\varphi$:  
\begin{equation}
\begin{gathered}
\label{Equation.6}
L_{f,\mathbf{E}}(\zeta,\psi) = \mathbb{E}_{x,a \sim \mathcal{D}} [\, \mathbb{E}_{\mathbf{z} \sim \mathbf{E}(\mathbf{z}|x_t,x_{t+1})}[\, f_{\zeta,\psi}(\mathbf{z}_g, \mathbf{z'}_h, \mathbf{z}_h)]\,]\, - \, \mathbb{E}_{x,a \sim \pi} [\, \mathbb{E}_{\mathbf{z} \sim \mathbf{E}(\mathbf{z}|x_t,x_{t+1})}[\, f_{\zeta,\psi}(\mathbf{z}_g, \mathbf{z'}_h, \mathbf{z}_h)]\,]
\\ +\, \varphi \, (\mathbb{E}_{x,a \sim \pi} \, [\,\text{KL} [\mathbf{E}(\mathbf{z}|x_t,x_{t+1})] \, || \, r(\mathbf{z}) \,] - I_c)
\end{gathered}
\end{equation}
where the mutual information between dialog states $X$ and latent variable $Z$ is
\begin{align*}
I(Z,X) &= \text{KL} [p(\mathbf{z},x) || p(\mathbf{z})p(x)] \\
&= \displaystyle\int d\mathbf{z} \ dx \ p(\mathbf{z},x) \log{\cfrac{p(\mathbf{z},x)}{p(\mathbf{z})p(x)}} = \displaystyle\int d\mathbf{z} \ dx \ p(x)\mathbf{E}(\mathbf{z}|x) \log{\cfrac{\mathbf{E}(\mathbf{z}|x)}{p(\mathbf{z})}} \\
&\le I_c = \displaystyle\int d\mathbf{z} \ dx \ \pi_{\theta}(x) \mathbf{E}(\mathbf{z}|x) \log{\cfrac{\mathbf{E}(\mathbf{z}|x)}{r(\mathbf{z})}} = \mathbb{E}_{x,a \sim \pi}[ \text{KL} [\mathbf{E}(\mathbf{z}|x) || r(\mathbf{z})] ] 
\end{align*}
In Equation \ref{Equation.6}, the VRB minimizes the mutual information with dialog states to focus on discriminative features. The VRB also minimizes the KL-divergence with the human dialogs, while maximizing the KL-divergence with the generated dialogs, thereby distinguishing effectively between samples from human dialogs and dialog policy. Our proposed model is summarized in Appendix \ref{Appendix B}.
\begin{figure}
    \centering
    \includegraphics[height=7cm, width=16cm]{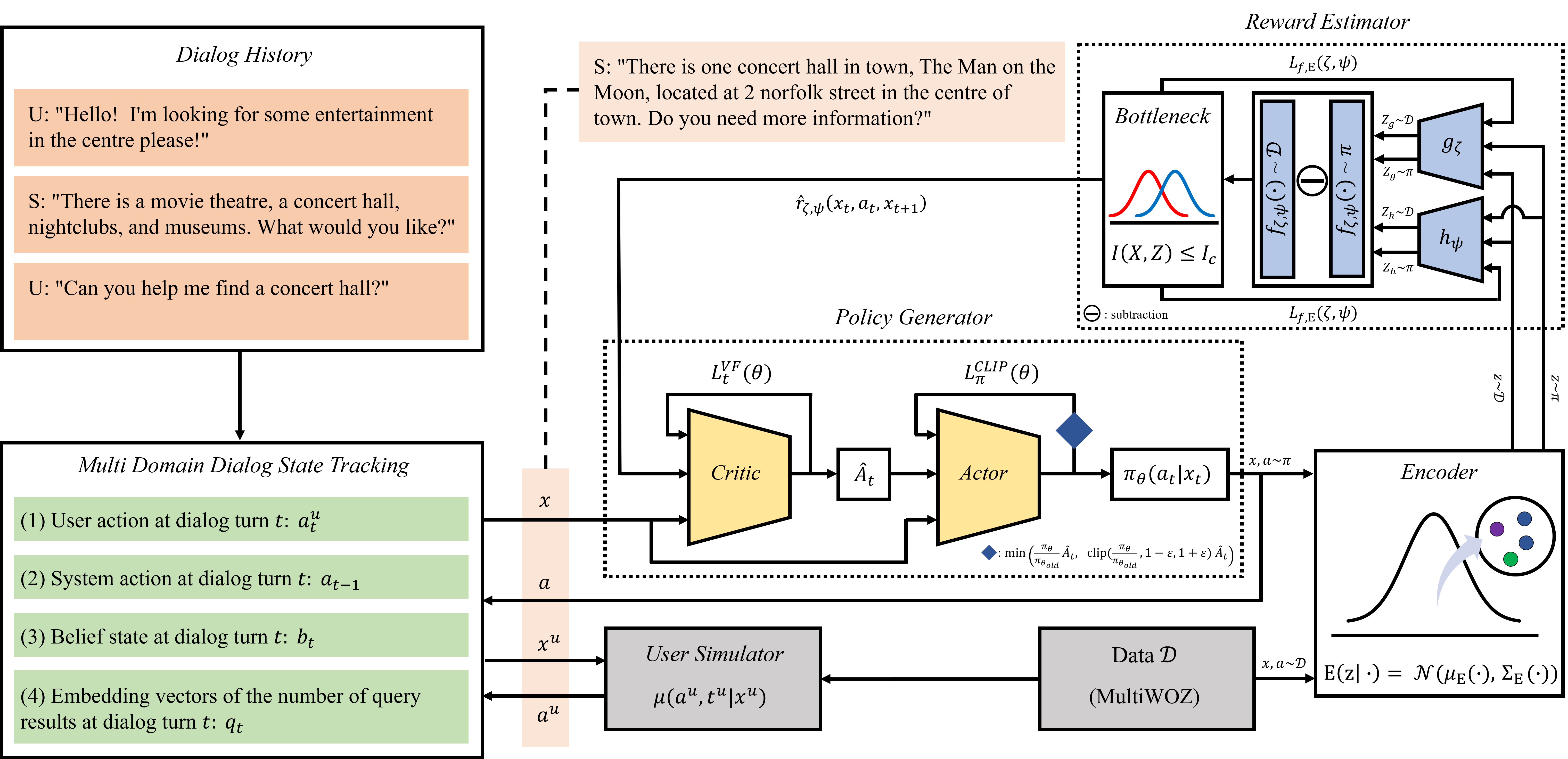}
    \caption{Schematic depiction of Variational Reward Estimator Bottleneck.}
    \label{fig:2}
\end{figure}
\section{Experiments}
\subsection{Dataset}
We evaluate our proposed method on Multi-domain wizard-of-oz \cite{budzianowski-etal-2018-multiwoz} (MultiWOZ), which contains approximately 10,000 of large-scale, multi-domain, and multi-turn conversational dialog corpora. MultiWOZ consists of seven distinct task-oriented domains, 24 slots, and 4,510 slot values. The dialog sessions are randomly divided into training, validation, and test set. The validation and test sets contain 1,000 sessions each.
\subsection{Training Details}
To demonstrate the robustness of our model, we conduct experiments over 30 times for each user simulator and average the results. We use the agenda-based user simulator \cite{schatzmann-etal-2007-agenda} and VHUS-based user simulator \cite{gur2018user}. The policy network $\pi_\theta$ and value network $V$ are MLPs with two hidden layers. $g_{\zeta}$ and $h_{\psi}$ are MPLs with one hidden layer each. We use the ReLu activation function and Adam optimizer for the MLPs. The hyper-parameters are presented in Appendix \ref{Appendix C}.
\subsection{Results}
We compare the proposed method with the following existing methods: GP-MBCM \cite{7404871}, ACER \cite{DBLP:conf/iclr/0001BHMMKF17}, PPO \cite{schulman2017proximal}, ALDM \cite{liu-lane-2018-adversarial}, and GDPL \cite{takanobu-etal-2019-guided}. Moreover, we evaluate our proposed model using four metrics: (i) \textit{Turns}: we record the average number of dialog turns between the dialog agent and user simulator. (ii) \textit{Match rate}: we conduct \textit{match rate} experiments to analyze whether the booked entities are matched with the corresponding constraints in the multi-domain environment. For instance, in Figure \ref{fig:2}, \textit{entertainment} should be matched with \textit{concert hall in the centre}. The match rate ranges from 0 to 1, and scores 0 if an agent fails to book the entity. (iii) \textit{Inform F1}: we test the ability of the model to inform all of the requested slot values. For example, in Figure \ref{fig:1}, the price range, food type, and area should be informed if the user wishes to visit a \textit{high-end} \textit{Cuban restaurant} in \textit{Cambridge}. (iv) \textit{Success rate}: in the \textit{success rate} experiment, a dialog session scores 0 or 1. We obtain 1 if all required information is presented and every entity is booked successfully. 

Table \ref{table1} presents the empirical results on both simulators and MultiWOZ. In the agenda-based setting, we observe that our proposed method achieves a new state-of-the-art performance. Note that an outstanding model should obtain high scores in every metric, not just a single one, because to regard a dialog as having ended successfully, every request should be informed precisely, thereby guiding a dialog through the user goal. Although GDPL achieves the highest score in Inform F1, our proposed model acts more human-like with respect to \textit{Turns}, and provides more accurate slot values and matched-entities than the other methods. In VHUS setting, on the other hand, though PPO behaves more human-like in \textit{Turns}, PPO exhibits greater difficulty in providing accurate information, while our model doesn't because our method constrains unproductive information flows. Both results in Table \ref{table1} demonstrate that our proposed model outperforms existing models, providing more definitive information than the other methods. 
\begin{table*}[t]\centering
\begin{adjustbox}{width=0.95\textwidth}
\begin{tabular}{l|cccc|cccc}   
\toprule                     
\multirow{2}{*}{Model} &\multicolumn{4}{c|}{Agenda} & \multicolumn{4}{c}{VHUS}  \\ 
& Turns & Match & Inform & Success & Turns & Match & Inform & Success \\
\midrule
GP-MBCM \cite{7404871} &2.99 &19.04 &44.29 &28.9 & - & - & - & - \\ 
ACER \cite{DBLP:conf/iclr/0001BHMMKF17} &10.49 &77.98 &62.83 &50.8 &22.35 &33.08 &55.13 &18.6 \\ 
PPO \cite{schulman2017proximal} &9.83 &83.34 &69.09 &59.1 &\textbf{19.23} &33.08 &56.31 &18.3 \\ 
ALDM \cite{liu-lane-2018-adversarial} &12.47 &81.20 &62.60 &61.2 &26.90 &24.15 &54.37 &16.4 \\
GDPL \cite{takanobu-etal-2019-guided} &7.64 &83.90 &\textbf{94.97} &86.5 &22.43 &36.21 &52.58 &19.7 \\
\midrule
\textbf{VRB (Ours)}  &\textbf{7.59} &\textbf{90.87} &90.97 &\textbf{90.4} &20.96 &\textbf{44.93} & \textbf{56.93} & \textbf{20.1} \\
\midrule
\textit{Human} &7.37 &95.29 &66.89 &75.0 &- &- &- &- \\
\bottomrule
\end{tabular}
\end{adjustbox}
\caption{Results of Agenda-based and VHUS-based user simulators.}
\label{table1}
\end{table*}
\section{Conclusions}
In this paper, we develop a novel and effective regularization method known as the Variational reward estimator bottleneck (VRB) for multi-domain task-oriented dialog systems. VRB contains a stochastic encoder which enables the reward estimator to be maximally informative, as well as provides information bottleneck regularization, which constrains unproductive information flows between the inputs and reward estimator. The empirical results demonstrate that VRB achieves a new state-of-the-art performances on two different user simulators and a multi-turn and multi-domain task-oriented dialog dataset.
\clearpage

\bibliographystyle{coling}
\bibliography{coling2020}

\clearpage
\appendix
\section*{Appendix}
\section{Mathematical Details}
\label{Appendix A}
\subsection{Background and Notations on MDP}
\label{app a.1}
To represent Inverse reinforcement learning (IRL) as a Markov decision process (MDP), we consider a tuple $\mathcal{M}$ =  $(\mathcal{X},\mathcal{A},T,\mathcal{R}, \rho_0, \gamma)$, where $\mathcal{X}$ is state space and $\mathcal{A}$ is the action space. The transition probability $T(x_{t+1}|x_{t},a_{t})$ defines the distribution of the next state $x_{t+1}$ given state $x_{t}$ and $a_{t}$ at time-step t. $\mathcal{R}(x_t,a_t)$ is the reward function of the state-action pair, $\rho_{0}$ is the distribution of the initial state $x_0$, and $\gamma$ is the discount factor. The stochastic policy $\pi(a_t|x_t)$ maps a state to a distribution over actions. Supposing we are given an optimal policy $\pi^*$, the goal of IRL is to estimate the reward function $\mathcal{R}$ from the trajectory $\tau = \{x_0,a_0,x_1,a_1, ..., x_T,a_T\} \sim \pi^*$. However, constructing an effective reward function is challenging, especially in multi-domain task-oriented dialog system.
\subsection{Gradient-Based Optimization}
\label{app a.2}
To imitate human behaviors, the reward estimator should learn the distributions of human dialog sessions using the KL-divergence loss:
\begin{align*}
L_{\pi}(\theta) &\approx -\text{KL} \left( \pi_{\theta}(\tau) \ || \ \cfrac{\exp(\mathcal{R_{\zeta})}}{Z} \right) \\
&= \sum{\pi_{\theta}(\tau) \ \log{\left( \cfrac{\cfrac{\exp(\mathcal{R_{\zeta})}}{Z}}{\cfrac{\pi_{\theta}(\tau)}{1}} \right)}} \\ 
&= \mathbb{E}_{\tau \sim \pi}[\log{\left( \cfrac{\exp(\mathcal{R_{\zeta})}}{Z} \right) } - \log{\pi_{\theta}(\tau)}]\\ 
&= \mathbb{E}_{\tau \sim \pi}[f_{\zeta}(\tau) - \log{\pi_{\theta}(\tau)}] \\
&= \mathbb{E}_{x, a \sim \pi}[f_{\zeta,\psi}(x_t, a_t, x_{t+1})] + \mathbb{E}_{s, a \sim \pi}[- \log{\pi_{\theta}(x_t, a_t, x_{t+1})}] \\ 
&= \mathbb{E}_{x, a \sim \pi}[f_{\zeta,\psi}(x_t, a_t, x_{t+1})] + H(\pi_\theta) 
\end{align*}
where $H(\pi_\theta)$ is the entropy of dialog policy $\pi_\theta$. The reward estimator maximizes the entropy, which represents maximizing the likelihood of observed dialog sessions. Therefore, the reward estimator is trained to discern between human dialog sessions $\mathcal{D}$ and dialog sessions that are generated by the dialog policy:
\begin{align*}
L_{f}(\zeta, \psi) &=  - \text{KL} \left( \mathcal{D}(\tau) \ || \ \cfrac{\exp(\mathcal{R_{\zeta})}}{Z} \right) - \left( -\text{KL} \left( \pi_{\theta}(\tau) \ || \ \cfrac{\exp(\mathcal{R_{\zeta})}}{Z} \right) \right) \\
    &= \mathbb{E}_{x, a \sim \mathcal{D}}[f_{\zeta, \psi}(x_t, a_t, x_{t+1})] + H(\mathcal{D}) - \mathbb{E}_{s, a \sim \pi}[f_{\zeta, \psi}(x_t, a_t, x_{t+1})] \ - H(\pi_\theta)
\end{align*}
Note that $H(\mathcal{D})$ and $H(\pi_\theta)$ are not dependent on the parameters $\zeta$ and $\psi$. Thus, the reward estimator can be trained using gradient-based optimization as follows: 
\begin{align*}
L_{f}(\zeta,\psi) = \mathbb{E}_{x,a \sim \mathcal{D}}[f_{\zeta,\psi}(x_t, a_t, x_{t+1})] - \mathbb{E}_{x,a \sim \pi}[f_{\zeta,\psi}(x_t, a_t, x_{t+1})]
\end{align*}
\subsection{Discriminative Reward Function}
\label{app a.3}
The reward function $\hat{r}_{\zeta,\psi}$ can be simplified in the following manner:
\begin{align*}
\hat{r}_{\zeta,\psi}(x_t,a_t,x_{t+1}) &=  \log{[ D_{\zeta,\psi}(x_t,a_t,x_{t+1}) ]} - \log{[ 1 - D_{\zeta,\psi}(x_t,a_t,x_{t+1}) ]}\\ 
&= \log{\left[ -1 + \cfrac{1}{1-D_{\zeta,\psi}(x_t,a_t,x_{t+1})} \right]} \\
&= \log{\left[ \cfrac{\exp{[ f_{\zeta,\psi}(x_t,a_t,x_{t+1}) ]}}{\pi_\theta(a_t | x_t)} \right]}\\
&= f_{\zeta, \psi}(x_t,a_t,x_{t+1}) - \log{\pi}_{\theta}(a_t | x_t)
\end{align*}
\section{Algorithm}
\label{Appendix B}
\begin{algorithm}[h]
\DontPrintSemicolon
Initialize dialog policy generator $\pi_\theta$ and reward estimator $f_{\zeta,\psi}$ \\
\For{$i \gets 0$ \textbf{to} $N$}
{
    Obtain random samples from human dialog corpus $\mathcal{D}$ \\
    Gather dialog sessions using user simulator $\mu(a^u,t^u|x^u)$ and policy generator $\pi_\theta(a|x)$\\
    Encode dialog sessions using stochastic encoder $\mathbf{E}(\mathbf{z}|\cdot) = \mathcal{N}(\mu_{\mathbf{E}}(\cdot), \Sigma_{\mathbf{E}}(\cdot))$\\
    Compute information bottleneck $\mathbb{E}_{x,a \sim \pi}[ \text{KL} [\mathbf{E}(\mathbf{z}|x) || r(\mathbf{z})] ] $ \\
    Update reward estimator $f_{\zeta,\psi}$ by optimizing $L_{f,\mathbf{E}}(\zeta, \psi)$ (Equation \ref{Equation.6}) \\
    Estimate reward function $\hat{r}_{\zeta, \psi}$ for each state-action pair \\
    Update state-value function $V(\mathcal{X})$ and dialog policy $\pi_\theta$  given the reward $\hat{r}_{\zeta, \psi}$ (Equation \ref{Equation.4})
}
\caption{Variational Reward Estimator Bottleneck}
\end{algorithm}
\section{Hyperparameters}
\label{Appendix C}
\begin{table}[h]
\centering
\begin{tabular}{l|c}
\textbf{Hyperparameter} & \textbf{Value} \\ \hline
Lagrange multiplier $\varphi$ & 0.001 \\ \hline
Upper bound $I_c$ & 0.5 \\ \hline
Learning rate of dialog policy & 0.0001 \\ \hline 
Learning rate of reward estimator & 0.0001 \\ \hline
Learning rate of user simulator & 0.001  \\ \hline
Clipping component $\epsilon$ for dialog policy & 0.02  \\ \hline
GAE component $\lambda$ for dialog policy & 0.95  \\ \hline
\end{tabular}
\caption{VRB hyperparameters.}
\label{table2}
\vspace{-2ex}
\end{table}
\end{document}